\documentclass[journal,web]{article}
\usepackage{cite}
\usepackage{amsmath,amssymb,amsfonts}
\usepackage{algorithm2e}
\usepackage{graphicx}
\usepackage{textcomp}
\usepackage{array}
\usepackage{graphicx}
\usepackage{caption}
\usepackage{subcaption}
\usepackage{enumerate}
\usepackage{booktabs}
\usepackage{multirow}
\usepackage{threeparttable}
\usepackage{verbatim}
\usepackage{placeins}
\usepackage{flushend}
\usepackage{bbm}
\newcommand{\myparagraph}[1]{\vspace{0.1em}\noindent\textbf{#1}}
\usepackage[colorlinks,hyperindex,breaklinks]{hyperref}
\usepackage[noend]{algpseudocode}
\usepackage[font=small,labelfont=bf,justification=justified,format=plain]{caption} 
\newcommand{\ie}{\textit{i}.\textit{e}.}
\newcommand{\eg}{\textit{e}.\textit{g}.}
\newcommand{\etal}{\textit{et al}.}
\newlength\myindent
\def\BibTeX{{\rm B\kern-.05em{\sc i\kern-.025em b}\kern-.08em
		T\kern-.1667em\lower.7ex\hbox{E}\kern-.125emX}}
\begin{document}
	\title{FCA: Taming Long-tailed Federated Medical Image Classification by Classifier Anchoring}
	\author{Jeffry Wicaksana, Zengqiang Yan, and Kwang-Ting Cheng
		\thanks{
			Jeffry Wicaksana and Kwang-Ting Cheng are with the Department of Electronic and Computer Engineering, Hong Kong University of Science and Technology, Kowloon, Hong Kong (E-mail: jwicaksana@connect.ust.hk, timcheng@ust.hk).
		}
		\thanks{
			Zengqiang Yan is with the School of Electronic Information and Communications, Huazhong University of Science and Technology, Wuhan, China (E-mail: z\_yan@hust.edu.cn).
		}
	}
	\maketitle
	
	\begin{abstract} 
        Limited training data and severe class imbalance impose significant challenges to developing clinically robust deep learning models. 
        Federated learning (FL) addresses the former by enabling different medical clients to collaboratively train a deep model without sharing data. However, the class imbalance problem persists due to inter-client class distribution variations. To overcome this, we propose federated classifier anchoring (FCA) by adding a personalized classifier at each client to guide and debias the federated model through consistency learning. 
        Additionally, FCA debiases the federated classifier and each client's personalized classifier based on their respective class distributions, thus mitigating divergence. 
        With FCA, the federated feature extractor effectively learns discriminative features suitably globally for federation as well as locally for all participants. 
        In clinical practice, the federated model is expected to be both generalized, performing well across clients, and specialized, benefiting each individual client from collaboration. 
        According to this, we propose a novel evaluation metric to assess models' generalization and specialization performance globally on an aggregated public test set and locally at each client. Through comprehensive comparison and evaluation, FCA outperforms the state-of-the-art methods with large margins for federated long-tailed skin lesion classification and intracranial hemorrhage classification, making it a more feasible solution in clinical settings. The code is available at: https://github.com/Jwicaksana/FCA.
	\end{abstract}
	
	\section{Introduction} 
	\label{sec:introduction}
	Computer-aided diagnosis based on medical image content analysis is a valuable tool for assisting professionals in decision making and patient screening. In recent years, deep learning models have achieved impressive success in various tasks, including skin lesion classification~\cite{skin1,skin2} of dermoscopy images, intracranial hemorrhage~\cite{rsna} identification of CT images, and autism disorder prediction~\cite{sheller,sheller2} of FMRI images, \textit{etc}. However, training a robust deep learning model requires a large amount of annotated data, which is infeasible in clinical scenarios. In addition, medical data is often imbalanced in nature~\cite{ham10k,isic2017,bcn2000} due to the varying prevalence of diseases, making it more challenging to develop models with high accuracy and generalizability to unseen data.
	
\begin{figure}[t!]
		\centering
		\includegraphics[width=0.9 \columnwidth]{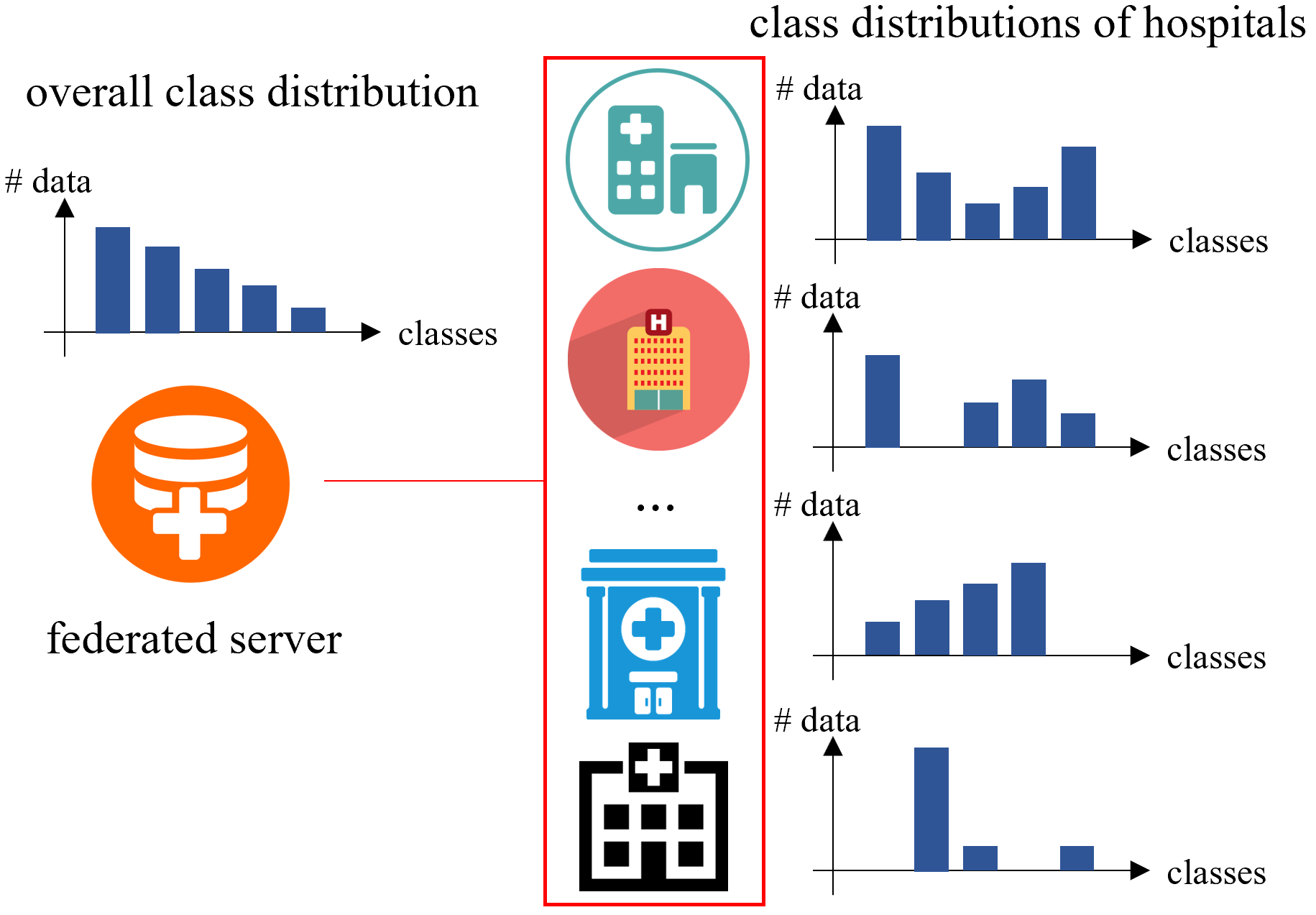}
		\caption{Variations of class distributions encountered under the federated setting. Variations occur between 1) different clients, where some clients have missing and rare classes, and 2) each individual client and the global federated server.} \label{class_statistics}
	\end{figure}

	Federated learning (FL) is a privacy-preserving solution that enables different medical clients to collaboratively train a federated model without sharing data~\cite{fedavg,sheller,sheller2,dou_npj}. In FL, a server facilitates collaboration by exchanging model weights instead of patients' data. A federated training round consists of two stages: 1) \textbf{local update}, where each client downloads the federated model from the server and updates it locally, and 2) \textbf{server update}, where the server aggregates model updates from each client and updates the federated model. The above federated training process repeats till convergence. Unfortunately, as shown in Fig.~\ref{class_statistics}, there exist class variations across clients, where each client's class distribution is not only imbalanced but also differs from the others. Such cross-client class imbalance can be fatal in FL, leading to unstable and slow training convergence~\cite{fedprox} and sub-optimal model performance~\cite{silo}.
	
	Classifier-guided learning, such as decoupling model learning into a feature extractor and a classifier~\cite{decouple}, provides a useful framework to better understand how to handle inter-client class variations in FL. Based on this, we categorize existing FL approaches into three categories:
	1) refining the federated classifier through model regularization\cite{fedprox,balancefl,moon},
	2) freezing the federated classifier with random initialization as a consistent guide~\cite{fedbabu}, and
	3) utilizing multiple personalized classifiers as guides~\cite{pros_nas,prr,cusfl,fedrep,fedbabu}.
	Unfortunately, none of these solutions is ideal. Model regularization forces different clients to conform to the average, which can be detrimental for clients with varying class distributions. Freezing classifiers with random initialization reduces divergence with a stable guide, but such a randomly initialized frozen guide may distort feature extractors. Using multiple personalized classifiers can better capture each client's distribution, but it can increase feature extractors' divergence as each client is optimized with a different target.
	
	In this paper, we present a Federated Classifier Anchoring (FCA) method, which is designed to train a robust federated feature extractor that can handle variations in clients' class distributions. It is achieved by combining multiple personalized classifiers with model regularization to address inter-client class variations. Specifically, in FCA, a personalized classifier is added as an anchor at each client to guide the federated model, and the federated feature extractor is trained with multiple experts including a generalization expert (such as the federated classifier) and multiple specialization experts (such as the personalized classifier of each client). To improve the robustness of extracted features, FCA first removes each client's bias through classifier calibration based on local client's class distributions and then imposes consistency regularization between the predicted logits of the federated classifier and each client's personalized classifier. As the personalized classifier is locally maintained, it provides more stable guidance compared to the federated classifier.
	
	In real-world scenarios, a federated model is expected to operate globally across multiple clients to achieve generalization and locally at each client to achieve specialization. For a more comprehensive evaluation, we propose to evaluate models' performance on both an aggregated public test set for generalization and a local test set at each client for specialization. Extensive experiments on real-world federated long-tailed datasets for skin lesion \cite{ham10k,isic2017,bcn2000} and intracranial hemorrhages classification ~\cite{rsna} demonstrate the superiority of FCA against the state-of-the-art approaches on both generalized and specialized evaluation settings. 
	
	\begin{figure*}[t!]
		\centering	\includegraphics[width=0.85\textwidth]{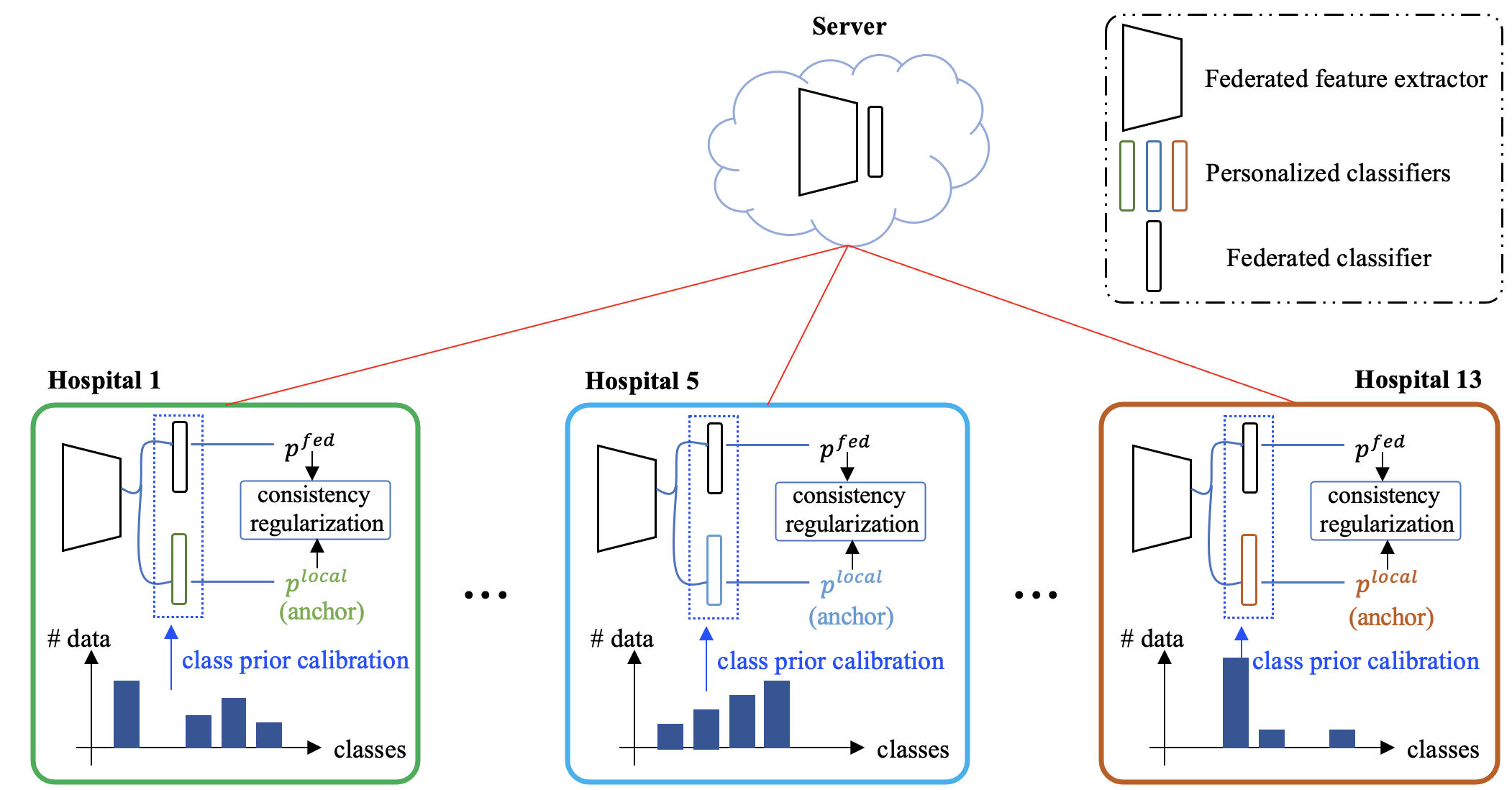}
		\caption{
			Illustration of the proposed federated classifier anchoring (FCA). FCA introduces a personalized classifier at each client as an anchor for the federated model by 1) debiasing classifiers according to each client's local class distribution and 2) guiding federated model updating through consistency regularization between classifiers. 
		}
		\label{overview}
	\end{figure*}
	
	Our contributions are summarized as follows:
	\begin{itemize}
		\item We view federated learning from the lens of classifier-guided learning and propose FCA to handle long-tailed federated learning for medical image classification. FCA leverages each client's personalized classifier to guide the federated model in learning more robust and discriminative features. 
		\item We propose a new evaluation metric to evaluate the generalization and specialization performance of federated learning solutions. Generalization is evaluated on an aggregated test set from clients in the federation while specialization is evaluated locally at each client. 
		\item We evaluate FCA on challenging long-tailed skin lesion classification and intracranial hemorrhage identification, where FCA consistently outperforms the state-of-the-art methods by large margins.
	\end{itemize}
	
	The rest of this paper is organized as follows. Related works are summarized and discussed in Section~\ref{lit}. Details of FCA are introduced in Section~\ref{framework}. In Section~\ref{evaluation}, we present a thorough evaluation of FCA compared with the existing methods and provide ablation studies as well as analysis in Section~\ref{discussion}. Section~\ref{conclusion} concludes the paper.
	
	\section{Related Work}\label{lit}
	
	\subsection{Federated Learning}
	
	Federated learning (FL)~\cite{fedavg} is desirable to the medical imaging communities~\cite{fl_natural, dou_npj} as it enables different clients to collaboratively train a deep model without sharing raw data. FL has been validated for multi-site functional magnetic resonance imaging classification~\cite{abide}, health tracking through wearables~\cite{fl_wearable}, COVID-19 screening and lesion detection~\cite{fl_covid}, brain tumor segmentation~\cite{fl_pp_brain, sheller,sheller2}, skin tumor classification~\cite{prr}, \textit{etc}. 
	
	In practice, each client collects its data locally under different conditions and protocols~\cite{fl_dichotomy}, usually resulting in inter-client statistical variations. To learn a robust model, inter-client statistical heterogeneity~\cite{flamby,fedavg,flc} (\textit{i.e.}, non-IID) must be addressed. Furthermore, a federated model should perform well for every participant in the federation, \ie, both generalization and specialization. However, existing approaches for handling inter-client variations tend to focus on either improving generalization over all clients or specialization through multiple personalized models, rather than addressing both simultaneously.
	
	Approaches focusing on generalization aim to reduce the divergence between the global federated model and the local model updates of each client. FedProx~\cite{fedprox} and FedMA~\cite{fedma} minimized the divergence in model weights with a proximal term and constructed the global federated model in a layer-wise manner respectively. 
	FedDG~\cite{feddg}, VAFL~\cite{vafl}, and FedRobust~\cite{fedrobust} minimized variations in the sample space by mapping each client's data into a common domain. In the feature space, FTL~\cite{eegdomain} regularized features across clients with a covariance matrix, MOON~\cite{moon} and FedCON~\cite{fedcon} used contrastive learning and FedBN~\cite{fedbn} leveraged local batch normalization. CCVR~\cite{ccvr} argued that classifiers diverge the most and thus debiased each client's classifier with virtually generated features. 
	
	Specialization-focused FL aims to maintain each client's unique characteristics and variations locally. FedRep~\cite{fedrep} and CusFL~\cite{cusfl} trained a federated feature extractor and allowed each client to keep its personalized classifier head. FedBABU~\cite{fedbabu} first froze the classifier during training and then fine-tuned it locally. FedMD~\cite{fedmd} and PRR~\cite{prr} trained different network architectures for each client with knowledge transfer, while \cite{pros_nas} utilized neural architecture search and automated machine learning. 
	
	In clinical practice, both generalization and specialization are important. However, improving generalization often is at the cost of degrading specialization, and vice versa. Motivated by this, FCA is designed to simultaneously achieve desirable generalization and specialization.
	
	\subsection{Class imbalance learning}
	
	Long-tailed centralized learning deploys: 1) data re-sampling, 2) data re-weighting, 3) representation learning, or 4) multi-expert learning. Data re-sampling either oversamples low-frequency classes~\cite{relay, calib_lt} or undersamples high-frequency classes~\cite{rethink_lt,systematic_imbalance}. Focal loss~\cite{focal_loss} and class-balanced loss~\cite{cbl} adjust the weight of each sample according to the training losses' magnitude and class frequencies respectively. Representation learning improves feature separation of different classes by increasing class margins~\cite{ldam}, learning class prototypical embeddings~\cite{paco}, calibrating classifiers to be class-balanced~\cite{balanced_softmax}, or training models in two stages~\cite{decouple}, \eg, representation learning and classifier learning. Multi-expert learning~\cite{bbn,trustworthy,sade} trains multiple classifier heads to handle different distributions, which is more practical in real applications.  
	
	Federated class imbalance learning is more challenging due to inter-client class variations. RatioLoss~\cite{fl_ci_sharing} estimated and reweighted each class' importance by monitoring model gradients in the server using auxiliary data. Astraea~\cite{astra} introduced a mediator as an oracle with access to rebalance each client's class distribution. CReRF~\cite{crerf} recalibrated the federated classifier in the server using synthesized balanced virtual features. FedIRM~\cite{fedirm} and imFedSemi~\cite{dynamic_bank} resolved class imbalance in semi-supervised settings by sharing class relation matrices and highly confident unlabeled samples respectively. Additional information sharing beyond model updates is not desirable in medical domains due to the potential risk of data leakage~\cite{leakage}. 
	To avoid sharing additional information, CLIMB~\cite{climb}, BalanceFL~\cite{balancefl}, and FedLC~\cite{fedlc} learned a balanced federated model by reweighting each client's importance during aggregation based on the empirical loss, balanced class sampling with self-entropy regularization, and logits calibration with pair-wise margins respectively. FedRS~\cite{fedrs} limited classifier updates when there are missing classes. However, training the federated model to conform to a balanced class distribution, may be detrimental for some clients, \eg, being sub-optimal compared to their locally-trained models. FCA overcomes the limitations by leveraging each client's personalized classifier as an anchor to guide and calibrate the global federated classifier and for local inference.
	
	\section{Methodology}\label{framework} 
	
	In this section, we first introduce notations in Section~\ref{sec3:1} and then provide an overview of federated learning with classifier anchoring (FCA), in Section~\ref{sec3:2}. Implementation details of FCA are presented in Sections~\ref{sec3:4} and \ref{sec3:5}. Finally, we analyze the intuition behind FCA from the classifier-guided representation learning perspective in Section~\ref{sec3:6}.
	
	\subsection{Preliminaries}\label{sec3:1}
	
	Federated model $\phi_{w}$ is denoted as a combination of feature extractor $f_{u}$ and classifier $g_{v}$, \eg, $\phi_{w} = \{f_{u}, g_{v}\}$, and optimized over $K$ clients' training data $D \triangleq \bigcup_{k} D_{k}$. Here, $D_{k}$ represents each client $k$'s local training data containing a long-tail data distribution with $C$ classes, namely $\{x_{i}, y_{i}\}$ where ${i}\in \{1, ..., |D_k|\}$ and $y_{i}\in \{1, ..., C\}$. In addition to $f_{u}$ and $g_{v}$, each client $k$ owns a locally-kept personalized classifier head $g_{v_{k}}$ as an anchor to guide the federated head $g_{v}$, and both $g_{v_{k}}$ and $g_{v}$ share the same feature extractor $f_{u}$.
	
	\subsection{Overview}\label{sec3:2}
	
	FCA resolves inter-client class variations by:
	\begin{enumerate}
		\item \textbf{Debiasing classifiers $g_{v}$ and $g_{v_{k}}$ according to each client's class distribution.} Debiasing enables both classifiers to pay equal attention to every class and reduces classifier divergence. 
		\item \textbf{Guiding and regularizing $g_{v}$ with $g_{v_{k}}$}. We use consistency learning~\cite{fixmatch} and view $g_{v_{k}}$ as a slightly perturbed version of $g_{v}$. Through penalizing prediction consistency between the two using knowledge distillation~\cite{icml19_kd}, extracted features of $f_{u}$ are encouraged to work well both locally, \eg, for $g_{v_{k}}$, and globally for $g_{v}$.   
	\end{enumerate}
	
	Each training round of FCA consists of two stages: 1) \textit{local client update}, where each client $k$ downloads the federated model $\phi_{w}$ from the server and updates both $\phi_{w}$ and locally kept $g_{v_{k}}$ with its local data, and 2) \textit{server update}, where the server updates the federated model with local updates from participating clients $\overline{\nabla}\phi= \{\nabla\phi_{w}^{1}, ..., \nabla\phi_{w}^{k} \}$. The pseudo-code of FCA is stated in Algorithm~\ref{alg}.
	
	\RestyleAlgo{ruled}
	\SetKwComment{Comment}{/* }{ */}
	\begin{algorithm}[t]
		\caption{Pseudocode of FCA}\label{alg}
		\SetKwInOut{Input}{input}
		\SetKwInOut{Output}{output}
		\SetKwInOut{Parameter}{parameter}
		\Input{$D_{i}$: Training data of each client $i$}
		\Parameter{$\lambda_{1}$, $\lambda_{2}$: balancing hyper-parameters of federated and personalized classifiers\\
			$\alpha$ : learning rate \\
			$T$: maximum federated training rounds\\}
		\Output{$w^{T}$: federated model's parameters\\ 
			$v_{1}^{T},..,v_{k}^{T}$: $K$ personalized classifiers' \\parameters}
		$w^{0}$, $v_{1}^{0},..,v_{k}^{0}$ $\leftarrow$ \textbf{initialize}()\\
		\For{$t=1:T$}{
			${\overline{\nabla}\phi} = \{\}$\\
			\For{$i=1:K$}{
				$\phi_{w}$ $\leftarrow$ \textbf{Download}($w^{t-1}$) \\
				$g_{v_{i}}$ $\leftarrow$ $v^{t-1}_{i}$ \\
				$\nabla {\phi}^{i}_{w}$, $\nabla {v^t_{i}}$ $\leftarrow$ \textbf{Update}($\phi_{w}, g_{v_{i}}; \lambda_{1}, \lambda_{2},D_i$)\\
				$v^t_{i} \leftarrow v^{t-1}_{i} - \alpha * \nabla {v^t_{i}}$\\
				$\overline{\nabla}\phi$.add($\nabla{\phi}^{i}_{w}$) \\
			}
			${w}^{t}$ $\leftarrow$ \textbf{Aggregate}(${\overline{\nabla}\phi}$, $w^{t-1}$)\\
		}
		\Return{$w^{T}$, $v_{1}^{T}$ , .., $v_{k}^{T}$}
	\end{algorithm}
	
	\subsection{Local client update}\label{sec3:4}
	
	Local client update consists of classifier calibration and consistency regularization. 
	For brevity, given any input $x_i$, we denote the logits predictions of both federated classifier and personalized classifier as $p^{fed}_{i}$ and $p^{local}_{i}$ respectively, calculated by
	\begin{equation}\label{fed}
	\begin{split}
	p^{fed}_{i} = g_{v}(f_{u}(x_{i})),\\
	p^{local}_{i} = g_{v_k}(f_u(x_{i})).
	\end{split} 
	\end{equation}
	
	\subsubsection{Classifier calibration}
	
	For calibration, we adopt balanced softmax loss~\cite{balanced_softmax} which uses class frequencies as a prior to compensate long-tailed class distributions. Let $\pi_{k} = [\pi^{0}_{k}, ... , \pi^{|class|}_{k}]$ and $\pi^{c}_{k} = |D^c_k|/|D_k|$ be the frequency of class $c$ at client $k$, where $D^c_k=\sum_{x_{i}\in D_{k}} \mathbbm{1}_{y_i=c}$. 
	Both federated and personalized classifiers are to optimize the following losses for calibration,
	\begin{equation}\label{lfed}
	\begin{split}
	\mathcal{L}_{k}^{fed} = \dfrac{1}{n_{k}} \sum_{x_{i}\in D_{k}} -y_{i}\text{log }\sigma (p^{fed}_{i} +\text{log}\pi_k),\\
	\mathcal{L}_{k}^{local} = \dfrac{1}{n_{k}} \sum_{x_{i}\in D_{k}} -y_{i}\text{log }\sigma (p^{local}_{i}  +\text{log}\pi_k),
	\end{split}
	\end{equation}
	where $\sigma(.)$ is the softmax function.
	
	\subsubsection{Classifier anchoring}
	
	As each client's personalized classifier is locally-kept and not disrupted by federated averaging, it is better at capturing each client's distribution. Thus, we utilize $p_{i}^{local}$ to guide $p_{i}^{fed}$ for more effective specialization and stop gradient flow to $g_{v_{k}}$, \eg, $\not\rightarrow$ indicates a stop gradient. The optimization graph is formed as:
	\begin{equation}
	\begin{split}
	x_{i} \rightarrow f_{u} \rightarrow g_{v} \rightarrow p_{i}^{fed}\\
	\searrow g_{v_{k}} \not\rightarrow p_{i}^{local} 
	\end{split}
	\end{equation}
	For consistency regularization between $p_{i}^{fed}$ and $p_{i}^{local}$ of client $k$, a Kullback-Leibler (KL) divergence loss $\mathcal{L}^{con}_{k}$ is penalized to optimize the federated model $\phi_{w}$ by
	\begin{equation}\label{lcon}
	\mathcal{L}^{con}_{k} = \dfrac{1}{n_{k}} \sum_{x_{i}\in D_{k}} L_{KL}(p_{i}^{fed}, p_{i}^{local}).
	\end{equation}
	Therefore, the overall loss of client $k$ during each local training round is
	\begin{equation}
	\mathcal{L}_{k} = \lambda_{1}*\mathcal{L}_{k}^{fed} + \lambda_{2}*\mathcal{L}_{k}^{local} + \mathcal{L}_{k}^{con},
	\end{equation}
	where $\lambda_{1}$ and $\lambda_{2}$ are hyper-parameters to determine the importance of federated and personalized classifiers respectively.
	
	\subsection{Federated Model Update}\label{sec3:5}
	
	Federated averaging (FedAvg)~\cite{fedavg} is used to update the federated model in the server. Each client $k$ sends its local model update $\nabla \phi_{w}^{k}$ to the server and keeps its personalized head $g_{v_{k}}$ locally. Each client's importance weight is assigned according to its data amount and the server updates the federated model by
	\begin{equation}\label{fedavg}
	\phi_{w} \leftarrow \phi_{w} + \sum_{i=1}^{K} w_{i} \nabla \phi_{w}^{i},
	\end{equation}
	where $w_{k}=|D_k|/\sum_{i=1}^{K} |D_k|$.
	
	\subsection{Intuition Behind FCA}\label{sec3:6}
	
	Based on classifier-guided representation learning, we re-categorize existing FL approaches into the following:
	\begin{enumerate}
		\item \textbf{Regularizing classifiers' divergence.} Several methods penalize weight update~\cite{fedprox,fedma}, regularize input features~\cite{moon,fedcon,fedlc}, or calibrate guide according to a balanced class distribution~\cite{ccvr, fl_ci_sharing}. 
		Regularization can be viewed as anchoring different clients' local updates to the average, \eg, the federated model. However, minority clients, with different data distributions from the average, would actually suffer from regularization.  
		
		\item \textbf{Freezing classifier.} In this way, each client trains under a fixed criteria~\cite{fedbabu} which is more stable and in turn reduces local divergence. Unfortunately, random initialization may lead to a sub-optimal feature extractor.
		
		\item \textbf{Deploying multiple personalized classifiers.} Each client trains its personalized guide~\cite{fedrep,cusfl} while sharing the feature extractor. Personalization provides flexibility, as the federated feature extractor is encouraged to learn more generic features under multiple guides. However, satisfying multiple guides without explicit regularization between clients is difficult. Consequently, local updates may distort the federated feature extractor's robustness. 
	\end{enumerate}
	
	FCA can be viewed as a combination of multiple personalized classifiers and explicit classifier regularization. Specifically, personalized classifiers are used as stable anchors to guide the federated model with consistency regularization. Personalized classifiers are kept locally and more effective in capture each client's class distribution to provide more stable guidance, unlike the federated classifier, which is replaced and updated at each training round. 
	In terms of consistency regularization, in FCA, it is refined through the removal of clients' bias by calibrating both the federated and each client's personalized classifier according to the local class distribution. 
	It should be noted that the regularization from the federated classifier is utilized implicitly in FCA. As the federated classifier serves as a common guide for different clients, which restricts the divergence of each client's local updates, the feature extractor is encouraged to extract features that simultaneously generalize and specialize. 
	{\renewcommand{\arraystretch}{1.2}
		\begin{table}[t]
			\centering
			\caption{Statistics of different clients from the Fed-ISIC2019 dataset\cite{flamby}, preprocessed and curated from ISIC2019\cite{ham10k,isic2017,bcn2000}, including classes 0 (Melanoma), 1 (Melanocytic Nevus), 2 (Basal Cell Carcinoma), 3 (Actinic Keratosis), 4 (Benign Keratosis), 5 (Dermatofibroma), 6 (Vascular Lesion), and 7 (Squamous Cell Carcinoma).}\label{skin_info}
			\begin{tabular}{c|c|c|c|c|c}
				\hline
				\multirow{3}{*}{Source}  & \multicolumn{4}{c|}{ \# Images per class} & \# Images\\ \cline{2-6}
				& 0    & 1    & 2    & 3    & Train \\ \cline{6-6}
				& 4    & 5    & 6    & 7    & Test \\ \hline
				\multirow{2}{*}{Rosendahl}     & 342  & 803  & 296  & 109  & 1807 \\ \cline{6-6} 
				& 490  & 30   & 3    & 18   & 452     \\ \hline
				\multirow{2}{*}{BCN}           & 2857 & 4206 & 2809 & 737  & 9930 \\ \cline{6-6} 
				& 1138 & 124  & 111  & 431  & 2483     \\ \hline
				\multirow{2}{*}{MSK4}          & 215  & 415  & 0    & 0    & 655 \\ \cline{6-6} 
				& 189  & 0    & 0    & 0    & 164     \\ \hline
				\multirow{2}{*}{VIDIR Modern}  & 680  & 1832 & 211  & 21   & 2691 \\ \cline{6-6} 
				& 475  & 51   & 82   & 11   & 672     \\ \hline
				\multirow{2}{*}{VIDIR Old}     & 67   & 350  & 5    & 0    & 351 \\ \cline{6-6} 
				& 10   & 4    & 3    & 0    & 88   \\ \hline
				\multirow{2}{*}{VIDIR Molemax} & 24   & 3720 & 2    & 0    & 3163 \\ \cline{6-6} 
				& 124  & 30   & 43   & 0    & 791     \\ \hline
			\end{tabular}
		\end{table}
	}
	{\renewcommand{\arraystretch}{1.2}
		\begin{table}[t]
			\centering
			\caption{Statistics of different clients for federated intracranial CT hemorrhage detection~\cite{rsna} under the 5-client setting with mild inter-client class variations, including classes 0 (Epidural), 1 (Intraparenchymal), 2 (Intraventricular), 3 (Subarachnoid), and 4 (Subdural).}\label{brain_info}
			\begin{tabular}{c|c|c|c|c|c|c|c}
				\hline
				\multirow{2}{*}{Site}    & \multicolumn{5}{c|}{ \# Images per class} & \multicolumn{2}{c}{ \# Images} \\ \cline{2-8}
				& 0   & 1    & 2    & 3    & 4    & Train    & Test \\ \hline
				\#1 & 165 & 2109 & 1200 & 2373 & 4454 & 10301    & 2576     \\ \hline
				\#2 & 171 & 2209 & 1614 & 2148 & 4321 & 10463    & 2616     \\ \hline
				\#3 & 594 & 2474 & 1447 & 2220 & 5557 & 12292    & 3074     \\ \hline
				\#4 & 202 & 2282 & 1284 & 2576 & 4000 & 10344    & 2587     \\ \hline
				\#5 & 66  & 2071 & 1586 & 2456 & 4793 & 10972    & 2744     \\ \hline
			\end{tabular}
		\end{table}
	}
	{\renewcommand{\arraystretch}{1.2}
		\begin{table}[t]
			\centering
			\caption{Statistics of different clients for federated intracranial CT hemorrhage detection~\cite{rsna} under the 10-client setting with severely imbalanced inter-client class variations and missing classes, including classes 0 (Epidural), 1 (Intraparenchymal), 2 (Intraventricular), 3 (Subarachnoid), and 4 (Subdural).}\label{brain_info_split2}
			\begin{tabular}{c|c|c|c|c|c|c|c}
				\hline
				\multirow{2}{*}{Site}    & \multicolumn{5}{c|}{ \# Images per class} & \multicolumn{2}{c}{ \# Images} \\ \cline{2-8}
				& 0   & 1    & 2    & 3    & 4    & Train & Test \\ \hline
				\#1  & 0   & 0    & 1112 & 2245 & 4033 & 7390  & 1848     \\ \hline
				\#2  & 88  & 5522 & 0    & 1926 & 0    & 7536  & 1884     \\ \hline
				\#3  & 89  & 1122 & 2339 & 0    & 7627 & 11177 & 2795     \\ \hline
				\#4  & 0   & 2430 & 0    & 3574 & 2944 & 8948  & 2237     \\ \hline
				\#5  & 132 & 518  & 0    & 930  & 1828 & 852   & 2744     \\ \hline
				\#6  & 269 & 0    & 0    & 1034 & 1834 & 3137  & 785     \\ \hline
				\#7  & 538 & 928  & 2662 & 0    & 0    & 1033  & 2616     \\ \hline
				\#8  & 9   & 502  & 0    & 0    & 2869 & 3380  & 845     \\ \hline
				\#9  & 69  & 62   & 0    & 1635 & 957  & 2723  & 681     \\ \hline
				\#10 & 4   & 62   & 1017 & 429  & 1033 & 2545  & 637     \\ \hline
			\end{tabular}
		\end{table}
	}
	
	\section{Experiments}\label{evaluation}
	
	\subsection{Dataset and Preprocessing}
	
	Experiments are carried out on two challenging tasks:
	\begin{enumerate}
		
		\item \textbf{Skin lesion classification}. The Fed-ISIC2019~\cite{flamby} dataset contains 23,247 dermoscopy images from six medical sources including eight classes namely Melanoma, Melanocytic Nevus, Basal Cell Carcinoma, Actinic Keratosis, Benign Keratosis, Dermatofibroma, Vascular Lesion, and Squamous Cell Carcinoma. Each data source is regarded as a separate client. Statistical details of each client's data are presented in Table~\ref{skin_info}, where some clients, \eg, MSK4, ViDIR old, and ViDIR molemax, only contain partial classes. 
		\textbf{Preprocessing.} Following the recommendations in~\cite{kaggle_preproc}, each dermoscopy image is pre-processed with brightness normalization and color constancy and resized to 224$\times$224 pixels.
		
		\item \textbf{Intracranial Hemorrhage (ICH) Classification}. 
		The RSNA-ICH~\cite{rsna} dataset consists of CT images from four different medical sources with five sub-classes including Epidural, Intraparenchymal, Intraventricular, Subarachnoid, and Subdural.
		As data sources are not publicly available, we artificially split the data into two different multi-client settings. \textbf{Preprocessing.} Following \cite{fedirm,dynamic_bank}, 67,969 CT images with single hemorrhage type are selected and resized to 128$\times$128 pixels for training and testing. 
		
	\end{enumerate}
	
	\subsection{Evaluation}
	
	{\renewcommand{\arraystretch}{1.2}
		\begin{table*}[t]
			\centering
			\caption{
				Quantitative results of different learning frameworks for federated long-tailed skin lesion classification under two evaluation settings, \textit{i.e.},  the specialized and generalized evaluation sets. Each learning framework is trained under five different seeds and both the average performance and standard deviation are reported. The best results are marked in \textbf{bold}. 
			}
			\label{skin_full}
			\resizebox{\linewidth}{!}{\begin{tabular}{l|cc|cc|cc}
				\hline
				\multirow{2}{*}{Method} &  \multicolumn{2}{c|}{Specialization (S) (\%)}  &  \multicolumn{2}{c|}{Generalization (G) (\%)}  
				&  \multicolumn{2}{c}{Average of G\&S (\%)}   \\ \cline{2-7} 
				& bACC & bAUC & bACC & bAUC & bACC & bAUC \\ \cline{1-7}
				\multicolumn{7}{c}{Federated Averaging} \\ \hline
				CrossEntropy~\cite{fedavg}              & 63.5$\pm$1.1 & 90.7$\pm$0.6 & 63.1$\pm$0.5 & 92.5$\pm$0.2 & 63.3$\pm$0.5 & 91.6$\pm$0.4 \\ 
				Focal~\cite{focal_loss}                 & 62.9$\pm$1.6 & 90.3$\pm$0.8 & 57.9$\pm$0.8 & 92.3$\pm$0.1 & 60.4$\pm$0.9 & 91.3$\pm$0.4 \\ 
				BalancedSoftmax~\cite{balanced_softmax} & \textbf{69.5}$\pm$0.6 &\textbf{91.3}$\pm$0.7 & \textbf{69.0}$\pm$1.0 &\textbf{93.2}$\pm$0.4 & \textbf{69.3}$\pm$0.4 & \textbf{92.1}$\pm$0.5 \\ \hline
				\multicolumn{7}{c}{Learning Frameworks (with balanced softmax~\cite{balanced_softmax})} \\ \hline
				Local Learning                          & 70.0$\pm$0.7 & 88.3$\pm$1.2 & 33.7$\pm$0.5 & 71.2$\pm$0.6 & 51.9$\pm$0.5 & 79.8$\pm$0.5 \\ 
				FedAvg~\cite{fedavg}                    & 69.5$\pm$0.6 & 91.3$\pm$0.7 & 69.0$\pm$1.0   & 93.2$\pm$0.4 & 69.3$\pm$0.4 & 92.1$\pm$0.5 \\ 
				FedProx (MLSys20)~\cite{fedprox}        & 70.0$\pm$1.1 & 90.2$\pm$0.2 & 69.7$\pm$0.4 & 92.9$\pm$0.4 & 69.9$\pm$1.0 & 91.5$\pm$0.9 \\ 
				MOON( CVPR21)~\cite{moon}               & 69.5$\pm$1.0 & 90.5$\pm$0.8 & 67.6$\pm$1.0 & 92.7$\pm$0.3 & 68.6$\pm$0.9 & 91.6$\pm$0.5 \\ 
				CReRF (IJCAI22)~\cite{crerf}            & 67.6$\pm$1.3 & 89.6$\pm$1.2 & 68.6$\pm$0.9 & 89.7$\pm$0.5 & 68.1$\pm$1.3 & 89.7$\pm$0.9 \\ 
				FedRS (KDD21)~\cite{fedrs}              & 70.2$\pm$2.2 & 90.3$\pm$0.3 & 69.4$\pm$0.4 & 93.2$\pm$0.1 & 69.8$\pm$1.1 & 91.8$\pm$0.2 \\ 
				FedLC (ICML22)~\cite{fedlc}             & 67.8$\pm$0.9 & 89.3$\pm$0.9 & 69.5$\pm$0.9 & 92.6$\pm$0.2 & 68.7$\pm$0.8 & 91.0$\pm$0.5 \\ 
				BalanceFL (IPSN22)~\cite{balancefl}     & 66.4$\pm$1.0 & 88.2$\pm$0.4 & 70.8$\pm$0.7 & 92.3$\pm$0.4 & 68.6$\pm$1.0 & 90.3$\pm$0.5 \\ 
				FedREP (ICML21)~\cite{fedrep}           & 72.6$\pm$0.8 & 90.8$\pm$0.4 & 69.2$\pm$0.5 & 93.1$\pm$0.3 & 70.9$\pm$0.7 & 91.9$\pm$0.3 \\ 
				FedBABU (ICLR22)~\cite{fedbabu}         & 72.1$\pm$1.6 & 90.5$\pm$1.5 & 68.4$\pm$0.8 & 92.0$\pm$0.3 & 70.3$\pm$1.1 & 91.3$\pm$0.3 \\ 
				FCA (ours)                              & \textbf{75.9}$\pm$0.6 & \textbf{92.5}$\pm$0.6 & \textbf{74.3}$\pm$0.4 & \textbf{94.8}$\pm$0.2 & \textbf{75.1}$\pm$0.4 & \textbf{93.7}$\pm$0.3 \\  \hline
			\end{tabular}}
		\end{table*}
	}
	For each dataset/source, we use 80\% for training and 20\% for testing while preserving the same class ratio/distribution. The average performance and standard deviation of different learning frameworks through five-fold cross validation are reported for comparison. 
	
	\myparagraph{Metric.}
	Balanced accuracy (bACC), average per class ACC, balanced area under the curve (bAUC), and average per class AUC are jointly used for evaluation.
	
	\myparagraph{Settings.}
	As discussed above, both specialization and generalization of different learning frameworks are evaluated. 
	For specialization, model performance is separately evaluated on each client's test set, $D^{k}_{test}$, and measured by average bACC and bAUC over clients, namely $\frac{1}{K}\sum_{k=1}^{K} \text{bACC}({D^{k}_{test}}$) and $\frac{1}{K}\sum_{k=1}^{K} \text{bAUC}({D^{k}_{test}}$). 
	For generalization, model performance is evaluated on an aggregated shared test set $\sum_{k=1}^{K} {D}^k_{test}$ from every client and measured by bACC and bAUC. For unbiased evaluation, we also report the average specialization and generalization performance.
	
	\myparagraph{FCA evaluation.} As FCA consists of a federated classifier and multiple personalized classifier heads, the predictions of the federated classifier is used for generalization evaluation, and each client's personalized head is used for specialization evaluation. 
	
	\subsection{Implementation Details}
	
	\myparagraph{Network architectures.} Following~\cite{skin1,skin2}, EfficientNet-B0~\cite{efficientnet} is used as the baseline model architecture. For FCA, we replace the output classifier layer with two parallel linear layers corresponding to the federated and each client's personalized classifier.
	
	\myparagraph{Comparison methods.} Four types of approaches are included for comparison, including 1) local learning where each client trains a model individually, 2) FedAvg~\cite{fedavg} as a baseline comparison, 3) regularization-based FL approaches FedProx~\cite{fedprox}, MOON~\cite{moon}, CReRF~\cite{crerf}, and the most-recent state-of-the-art approaches such as BalanceFL~\cite{balancefl}, FedRS~\cite{fedrs}, and FedLC~\cite{fedlc}, and 4) the state-of-the-art personalized FL approaches FedRep~\cite{fedrep} and FedBABU~\cite{fedbabu}.
	
	\myparagraph{Training Details}. EfficientNet-B0 is initialized with the pre-trained weights from ImageNet and trained for 80 federated rounds using an Adam optimizer~\cite{adam} with a learning rate of 1e-3, a weight decay of 5e-4, and a batch size of 64. We apply learning rate decay with a factor of 0.1 at round 60 and 70. In synchronous federated training, each client $k$ updates the modal locally for one epoch and sends local model updates to the server at every federated round. 
	During local training, training images are augmented by random rotation, horizontal and vertical flipping, adding gaussian blur, and applying normalization. 
	For testing, each testing image is normalized according to the training statistics. For a fair comparison, balanced softmax loss (BSM) \cite{balanced_softmax} is introduced to optimize all learning frameworks as it works better than regular cross entropy and focal loss~\cite{focal_loss} for federated long-tailed learning. 
	
	\subsection{Results on Skin Lesion Classification}
	
	\subsubsection{Experiment Settings}
	
	Fed-ISIC2019~\cite{flamby} is divided into six clients according to data sources: Rosendahl, BCN, MSK4, VIDIR Modern, VIDIR Old, and VIDIR Molemax respectively. As summarized in Table \ref{skin_info}, clients vary significantly in data amounts and class distributions, and there exist missing classes in MSK4, VIDIR Old, and VIDIR Molemax.
	
	\subsubsection{FedAvg with Inter-Client Class Variations}
	
	Three typical solutions to addressing long-tailed class imbalance, namely cross entropy loss, focal loss~\cite{focal_loss}, and balanced softmax~\cite{balanced_softmax}, are separately introduced to the baseline FedAvg~\cite{fedavg} for comparison as summarized in Table~\ref{skin_full}. On average, \eg, generalization and specialization performance, focal loss~\cite{focal_loss} underperforms cross entropy loss~\cite{fedavg} by an average of 2.9\% and 0.3\% in bACC and bAUC respectively. Though focal loss aims to up-weight hard samples, it may neglect the rare and missing classes. Comparatively, balanced softmax~\cite{balanced_softmax} can effectively debias predictions according to class distributions, outperforming cross entropy loss and focal loss evaluated on the average of generalization and specialization by 6.0\% and 0.5\% in bACC and bAUC respectively. Therefore, for the subsequent experiments, balanced softmax is introduced to all learning frameworks for a fair comparison.
	
	\subsubsection{Comparison of Various Learning Frameworks}
	
	As summarized in Table~\ref{skin_full}, FL generalizes better than local learning (LL), as it has access to a larger training set. However, FedAvg, MOON, CReRF, FedLC, and BalanceFL are sub-optimal compared to LL in bACC when evaluated on the specialization test set. In clinical practice, if collaborative learning is less beneficial compared to LL, it may disincentivize some clients from participating. 
	
	The state-of-the-art FL approaches included for comparison are categorized into three groups: 1) variation regularization based (FedPRox and MOON), 2) long-tailed focused (CReRF, BalanceFL, FedRS, and FedLC), and 3) personalized (FedBABU and FedREP).
	Though FedProx and MOON penalize the divergence between the federated model and local client updates, their performance is quite close to FedAvg. Specifically, MOON underperforms both FedAvg and FedProx under all evaluation settings as feature regularization with respect to a diverging federated model can be detrimental. It explains why regularization based on the federated model may not be helpful to debias clients' class variations. 
	While BalanceFL, FedRS, and FedLC outperform FedAvg on the generalized test set in bACC, they achieve sub-optimal performance in bAUC. It is because directly calibrating the classifier at each client according to its distribution without regularization may distort the model's decision boundaries. 
	On the specialized test set, FedRS slightly outperforms FedAvg in bACC by an average of 0.7\% as it decreases the weight updates on missing classes and makes clients focus only on existing classes. It should be noted that CReRF performs worse than other approaches as it relies on the generated features on the server to calibrate the classifier. When clients' local updates diverge, the quality of the generated features would be negatively affected. 
	{\renewcommand{\arraystretch}{1.2}
		\begin{table}[t]
			\centering
			\caption{
				Quantitative results of different learning frameworks for ICH classification under two splitting settings, \textit{i.e.,} \textbf{Split 1} (the 5-client setting containing imbalanced but no missing classes) and \textbf{Split 2} (the 10-client setting containing severely imbalanced with missing classes). We report the average bACC and bAUC on the generalized and specialized test sets. Each learning framework is trained under five different seeds and both the average performance and standard deviation are reported. The best results are marked in \textbf{bold}.  }
			\label{brain_full}
			\resizebox{\linewidth}{!}{
				\begin{tabular}{l|cc|cc}
					\hline
					\multirow{3}{*}{Method} &  \multicolumn{4}{c}{Average of Generalization \& Specialization (\%)} \\   \cline{2-5} 
					&  \multicolumn{2}{c|}{\textbf{Split 1}}    
					&  \multicolumn{2}{c}{\textbf{Split 2}}   \\ \cline{2-5} 
					& bACC & bAUC & bACC & bAUC \\ \cline{1-5}
					\multicolumn{5}{c}{Federated Averaging} \\ \hline
					CrossEntropy~\cite{fedavg}                &  62.5$\pm$0.3         & 91.5$\pm$0.1          & 53.7$\pm$0.7          & \textbf{88.4}$\pm0.3$ \\
					Focal~\cite{focal_loss}                   & 60.4$\pm$0.4          & 90.6$\pm$0.2          & 52.8$\pm$0.5          & 84.8$\pm$0.5  \\  
					BalancedSoftmax~\cite{balanced_softmax}   & \textbf{68.7}$\pm$0.4 & \textbf{92.2}$\pm$0.2 & \textbf{59.8}$\pm$0.4 & 84.8$\pm$0.2  \\ \hline
					\multicolumn{5}{c}{Learning Frameworks (with balanced softmax~\cite{balanced_softmax})} \\ \hline
					Local Learning                            & 61.3$\pm$0.1          & 87.7$\pm$0.1          & 52.4$\pm$0.4          & 77.8$\pm$0.2 \\ 
					FedAvg~\cite{fedavg}                      & 68.7$\pm$0.4          & 92.2$\pm$0.2          & 59.8$\pm$0.4          & 84.8$\pm$0.2 \\
					FedProx (MLSys 20)~\cite{fedprox}         & 69.0$\pm$0.4          & 91.7$\pm$0.2          & 60.1$\pm$0.3          & 88.7$\pm$0.5 \\ 
					MOON (CVPR21)~\cite{moon}                 & 67.8$\pm$0.5          & 91.0$\pm$0.3          & 59.1$\pm$0.4          & 88.1$\pm$0.5 \\ 
					CReRF (IJCAI 22)~\cite{crerf}             & 61.6$\pm$0.1          & 92.0$\pm$0.2          & 54.5$\pm$0.1          & 88.6$\pm$0.3 \\  
					FedRS (KDD21)~\cite{fedrs}                & 69.2$\pm$0.3          & 91.7$\pm$0.1          & 58.8$\pm$0.9          & 88.8$\pm$0.3 \\ 
					FedLC (ICML22)~\cite{fedlc}               & 69.0$\pm$0.3          & 91.6$\pm$0.1          & 57.4$\pm$0.3          & 85.6$\pm$0.5 \\ 
					BalanceFL (IPSN22)~\cite{balancefl}       & 74.0$\pm$0.1          & 92.6$\pm$0.1          & 60.8$\pm$0.7          & 88.5$\pm$0.3 \\ 
					FedREP (ICML21)~\cite{fedrep}             & 68.6$\pm$0.6          & 91.6$\pm$0.3          & 59.3$\pm$0.5          & 86.9$\pm$0.6 \\ 
					FedBABU (ICLR22)~\cite{fedbabu}           & 68.4$\pm$0.7          & 91.5$\pm$0.2          & 57.6$\pm$0.7          & 88.3$\pm$0.4 \\ 
					FCA (ours)                                & \textbf{75.6}$\pm$0.2 & \textbf{94.3}$\pm$0.1 & \textbf{66.2}$\pm$0.5 & \textbf{92.0}$\pm$1.0 \\  \hline
				\end{tabular}
			}
		\end{table}
	}
	{\renewcommand{\arraystretch}{1.2}
		\begin{table*}[t]
			\centering
			\caption{
				Ablation studies of FCA viewed from the lens of classifier-guided learning. The components are as follows:
				1) \textit{\# guide} indicates the number of classifiers used to train the federated feature extractor, 2) \textit{learnable guide} (\checkmark or $\times$) indicates whether guides/classifiers are learnable of frozen, and 3) \textit{regularization} (\checkmark or $\times$) indicates whether an explicit loss regularization is deployed during training. $K$ represents the total number of clients.
			}\label{ablation_cl}
			\resizebox{\linewidth}{!}{\begin{tabular}{ccc|cc|cc|cc}
				\hline
				\multicolumn{3}{c|}{\multirow{4}{*}{Components}}  
				& \multicolumn{6}{c}{Average of Generalization \& Specialization (\%)}   \\ \cline{4-9} 
				& & &  \multicolumn{2}{c|}{\multirow{2}{*}{Skin Lesion}} & \multicolumn{4}{c}{ICH} \\ \cline{6-9}
				& & & & &\multicolumn{2}{c|}{\textbf{Split 1}}& \multicolumn{2}{c}{\textbf{Split 2}}\\ \hline
				\# guide & learnable guide & regularization  & bACC         & bAUC         & bACC         & bAUC         & bACC         & bAUC \\ \hline
				1       & \checkmark      & $\times$        & 69.3$\pm$0.4 & 91.8$\pm$0.5 & 68.8$\pm$0.4 & 91.5$\pm$0.2 & 59.8$\pm$0.4 & 84.8$\pm$0.2\\ \hline  
				1       & \checkmark      & \checkmark      & 69.9$\pm$1.0 & 91.5$\pm$0.9 & 74.0$\pm$0.1 & 92.6$\pm$0.1 & 60.8$\pm$0.7 & 88.5$\pm$0.3  \\ \hline 
				$K$ & $\times$ & $\times$ & 70.3$\pm$1.1 & 91.3$\pm$0.3 & 68.4$\pm$0.4 & 91.5$\pm$0.2 & 59.3$\pm$0.5 & 86.9$\pm$0.6 \\ \hline 
				$K$ & \checkmark & $\times$ & 70.9$\pm$0.7 & 91.9$\pm$0.3 & 68.6$\pm$0.6 & 91.6$\pm$0.3 & 57.6$\pm$0.7 & 88.3$\pm$0.4 \\ \hline 
				$K+1$ & \checkmark & \checkmark & \textbf{75.1}$\pm$0.4 & \textbf{93.7}$\pm$0.3 & \textbf{75.6}$\pm$0.2 & \textbf{94.3}$\pm$0.1 & \textbf{66.2}$\pm$0.5 & \textbf{92.0}$\pm$0.1\\ \hline 
			\end{tabular}}
		\end{table*}
	}
	{\renewcommand{\arraystretch}{1.2}
		\begin{table*}[t]
			\centering
			\caption{
				Ablation studies of FCA on $\lambda_{1}$, the weight of the federated classifier's training loss, $\lambda_{2}$, the weight of each personalized classifier's training loss. \checkmark or $\times$ indicates the presence of consistency regularization (CR).
			}\label{ablation_lam}
			\resizebox{\linewidth}{!}{\begin{tabular}{cc|cc|cc|cc}
				\hline
				\multicolumn{2}{c|}{Components}  
				& \multicolumn{6}{c}{Average of Generalization \& Specialization (\%)}   \\ \hline 
				\multirow{5}{*}{$\lambda_{1}$} & \multirow{5}{*}{$\lambda_{2}$} &  \multicolumn{2}{c|}{\multirow{2}{*}{Skin Lesion (CR)}} & \multicolumn{4}{c}{ICH (CR)} \\ \cline{5-8}
				& & & &\multicolumn{2}{c|}{\textbf{Split 1}}& \multicolumn{2}{c}{\textbf{Split 2}}\\ \cline{3-8}
				& & $\times$ & \checkmark & $\times$ & \checkmark & $\times$ & \checkmark \\\cline{3-8} 
				&                                        &  bACC        & bACC         & bACC         & bACC         & bACC         & bACC \\ 
				&                                        & bAUC         & bAUC         & bAUC         & bAUC         & bAUC         &  bAUC      \\ \hline
				\multirow{2}{*}{1}  & \multirow{2}{*}{1} & 73.2$\pm$0.6 & 73.4$\pm$0.7 & 71.9$\pm$0.5 & 72.4$\pm$0.4 & 63.1$\pm$0.7 & 63.9$\pm$0.6\\
				&                    & 93.1$\pm$0.3 & 92.7$\pm$0.3 & 93.0$\pm$0.2 & 93.2$\pm$0.1 & 90.8$\pm$0.1 & 90.9$\pm$0.3\\ \hline
				\multirow{2}{*}{1}  & \multirow{2}{*}{2} & 73.7$\pm$0.6 & 74.3$\pm$0.8 & 73.8$\pm$0.4 & 74.0$\pm$0.4 & 63.5$\pm$0.2 & 63.8$\pm$0.4\\
				&                    & 93.0$\pm$0.4 & 92.9$\pm$0.4 & 93.8$\pm$0.1 & 93.8$\pm$0.1 & 91.0$\pm$0.1 & 91.5$\pm$0.2\\ \hline
				\multirow{2}{*}{1}  & \multirow{2}{*}{3} & 74.5$\pm$0.6 & \textbf{75.1}$\pm$0.4       & 74.6$\pm$0.2 & \textbf{75.6}$\pm$0.2 
				& 65.4$\pm$0.4 & \textbf{66.2}$\pm$0.5\\
				&                    & 93.0$\pm$0.5 & \textbf{93.7}$\pm$0.3       & 94.2$\pm$0.2 & \textbf{94.3}$\pm$0.1 
				& 91.9$\pm$0.1 & \textbf{92.0}$\pm$0.1\\ \hline
				\multirow{2}{*}{2}  & \multirow{2}{*}{1} & 72.2$\pm$0.9 & 72.9$\pm$0.6 & 73.3$\pm$0.4 & 73.7$\pm$0.3 & 64.8$\pm$0.7 & 65.0$\pm$0.1\\
				&                    & 92.3$\pm$0.5 & 92.9$\pm$0.4 & 93.5$\pm$0.1 & 93.7$\pm$0.2 & 91.6$\pm$0.4 & 91.6$\pm$0.2\\ \hline
				\multirow{2}{*}{3}  & \multirow{2}{*}{1} & 73.3$\pm$0.8 & 73.4$\pm$0.8 & 74.1$\pm$0.5 & 74.8$\pm$0.5 & 65.5$\pm$0.5 & 66.0$\pm$0.5\\
				&                    & 92.9$\pm$0.3 & 93.7$\pm$0.6 & 93.9$\pm$0.2 & 94.1$\pm$0.1 & 91.8$\pm$0.1 & 91.9$\pm$0.1 \\ \hline
			\end{tabular}}
		\end{table*}
	}
	
	FedRep and FedBABU significantly outperform single-model federated approaches on the specialized test set by an average increase of 2.4\% and 1.7\% respectively in bACC. Unfortunately, each client's personalized classifier overfits to its local distribution and fail to generalize, resulting in poor bACC performance on the generalized test set compared to BalanceFL and FedProx. 
	On the average of specialization and generalization, FedRep and FedBabu outperform single-model approaches by an average increase of 1.1\% and 0.4\% in bACC but underperforms FedAvg in bAUC by an average decrease of 0.3\% and 0.9\% respectively. Based on the bAUC results, their decision boundaries of different classes are less discriminative than FedAvg, highlighting the limitation of using multiple guides without regularization. 
	
	Comparatively, FCA outperforms the state-of-the-art federated and local learning approaches by an average increase of 3.3\% and 1.2\% in bACC and bAUC respectively on the specialized test set and 3.5\% and 1.6\% in bACC and bAUC respectively on the generalized test set. Through more consistent guidance provided by each client's debiased personalized classifier, FCA effectively learns a more robust federated model.
	
	\subsection{Results on Intracranial Hemorrhage Classification}
	
	\subsubsection{Experimental Settings.}
	
	Though the RSNA-ICH~\cite{rsna} dataset was collected from four different medical sources, the data source of each image is unknown. Therefore, following \cite{moon, crerf}, we use a Dirichlet distribution for data partitioning with cross-client class variations. Here, Dirichlet distribution is generated according to a hyper-parameter $\alpha$, where a higher $\alpha$ would lead to a more balanced distribution. Furthermore, to simulate missing classes, two settings are used for evaluation, including
	\begin{enumerate}
		
		\item \textbf{Split 1}: The 5-client FL setting with mild inter-client class variations. We utilize two different Dirichlet distributions according to class frequencies. In RSNA-ICH, Epidural is categorized as the minority class due to its limited data amount, \ie, 2.2\% of total data. Thus, we set $\alpha=0.5$ and $\alpha=50$ to distribute samples from the minority class, \ie, Epidural, and samples from the majority classes, \ie, the rest classes. Statistical details of different clients are summarized in Table~\ref{brain_info}. 
		
		\item \textbf{Split 2}: The 10-client FL setting with severe inter-client class variations and missing classes. We use five different Dirichlet distributions for classes, \ie, Subdural with $\alpha=50$, Subarachnoid with $\alpha=30$,  Intraventricular with $\alpha=10$, Intraparenchymal with $\alpha=5$, and Epidural with $\alpha=0.5$ respectively. To simulate missing classes, we randomly remove classes at each client with a probability of $0.3$. Statistical details are stated in Table~\ref{brain_info_split2}.
		
	\end{enumerate}
	
	\subsubsection{FedAvg with Inter-Client Class Variations}
	
	As summarized in Table~\ref{brain_full}, under \textbf{Split 1}, balanced softmax outperforms both cross entropy loss and focal loss by an average increase of 6.2\% and 0.7\% in bACC and bAUC respectively. Under \textbf{Split 2}, the federated model trained with cross entropy loss may over-emphasize samples of the majority classes, resulting in poor generalization. Consequently, though it achieves a higher bAUC compared to balanced softmax, the bACC performance is worse, suffering from an average decrease of 6.1\%. 
	
	\subsubsection{Comparison of Various Learning Frameworks}
	
	According to Table~\ref{brain_full}, FL consistently outperforms local learning (LL) in bACC and bAUC measured on the average of generalization and specialization test sets. Under \textbf{Split 1}, FedAvg and FedProx achieve comparable performance and even outperform most-recent state-of-the-art federated learning approaches like MOON, CReRF, FedRep, and FedBABU. 
	While calibrating logits according to each client's class distribution improves the performance of the federated model, \eg, FedRS and FedLC, it is insufficient to address severe inter-client class variations, leading to just slight improvement compared to FedProx. Comparatively, BalanceFL significantly outperforms FedProx by an average increase of 5.0\% in bACC and 0.4\% in bAUC, indicating that more attention should be placed to handle class imbalance and inter-client class variations.
	FCA outperforms BalanceFL by an average increase of 1.6\% in bACC and 1.7\% in bAUC respectively, demonstrating the effectiveness of consistency regularization and classifier debiasing in minimizing inter-client class variations and imbalance. 
	
	Under \textbf{Split 2}, BalanceFL outperforms FedProx, while both FedAvg and FedProx achieve comparable performance against most-recent state-of-the-art approaches. It should be noticed that FCA outperforms BalanceFL and FedLC by even larger margins, \eg, an average increase of 5.4\% and 6.3\% in bAUC respectively, validating the effectiveness of FCA in addressing more severe inter-client class variations.
	
	\section{Ablation Study}\label{discussion}
	
	\subsection{Revisit FCA from the Lens of Multi-Expert Learning}
	
	We first re-group various federated learning approaches into five categories according to the number of guides (classifiers) used, whether guides are learnable and whether regularization is used during training as summarized in Table~\ref{ablation_cl}. In Table~\ref{ablation_cl}, rows (from top to bottom) represent 1) FedAvg~\cite{fedavg}, 2) FL with regularization to reduce model update divergence in the weight space~\cite{fedprox}, the feature space~\cite{fedbn, moon}, the data space ~\cite{vafl, balancefl}, or classifiers~\cite{crerf,fedrs,fedlc}, 3) FedBABU which fixes the guide during training and fine-tune it to each client post federated training, 4) personalized federated learning~\cite{fedrep, cusfl} without explicit loss regularization during training, and 5) FCA which combines multiple personalized classifiers with explicit model regularization. Here, only the best-performing learning framework of each category is presented for comparison. 
	
	On skin lesion classification, training with multiple classifiers is more beneficial compared to relying on one single classifier as summarized in Table~\ref{ablation_cl}, indicating that having more experts could improve the quality of extracted features. 
	Combining multiple personalized classifiers with explicit regularization further improves the consistency with the guide and thus the overall performance. 
	On ICH classification, introducing regularization to the single-guide approach improves both bACC and bAUC by an average increase of 5.2\% and 0.9\% under \textbf{Split 1} and 1.0\% and 3.7\% under \textbf{Split 2} respectively. 
	Comparatively, training a federated solution based on multiple guides is sub-optimal compared to single-guide federated learning without regularization. 
	When the feature extractor fails to fit multiple and differing guides, the federated model suffers. Therefore, adding explicit regularization during optimization is essential to reduce the divergence between different clients' guides, especially when inter-client class variations are more severe, \eg, under \textbf{Split 2}. 
	FCA consistently outperforms other approaches in bACC and bAUC by an average increase of 1.6\% and 1.7\% under \textbf{Split 1} and 5.4\% and 3.5\% under \textbf{Split 2} respectively. The main improvements come from the multiple learnable guides with consistency regularization, effectively handling long-tailed federated learning. 
	
	\subsection{Hyperparameters of FCA}
	
	{\renewcommand{\arraystretch}{1.2}
		\begin{table}[t]
			\centering
			\caption{
				Ablation studies of FCA on the direction of consistency regularization (denoted as arrows). G and S are short for the evaluation on the generalization and specialization test sets respectively. 
			}\label{ablation_comp}
			\begin{tabular}{c|c|cc}
				\hline
				\multicolumn{1}{c|}{Components}                            & \multicolumn{3}{c}{Average of G\&S (\%)}   \\ \hline 
				\multirow{4}{*}{Regularization Direction}                  & \multirow{1}{*}{Skin Lesion} & \multicolumn{2}{c}{\multirow{1}{*}{ICH}} \\ \cline{3-4}
				&                              & \textbf{Split 1}      & \textbf{Split 2} \\ \cline{2-4}
				& bACC                         & bACC                  & bACC             \\ 
				& bAUC                         & bAUC                  & bAUC             \\ \hline
				\multirow{2}{*}{federated $\leftarrow$ personalized}       & \textbf{75.1}$\pm$0.4        & \textbf{75.6}$\pm$0.2 & \textbf{66.2}$\pm$0.5 \\
				& \textbf{93.7}$\pm$0.3        & \textbf{94.3}$\pm$0.1 & \textbf{92.0}$\pm$0.1 \\ \hline
				\multirow{2}{*}{federated $\rightarrow$ personalized}      & 74.0$\pm$0.2                 & 75.0$\pm$0.5          & 65.8$\pm$0.9  \\
				& 93.1$\pm$0.4                 & 94.2$\pm$0.1          & 91.9$\pm$0.2  \\ \hline
				\multirow{2}{*}{federated $\leftrightarrow$ personalized}  & 73.1$\pm$0.3                 & 74.5$\pm$0.4          & 64.4$\pm$0.8  \\
				& 92.9$\pm$0.4                 & 94.1$\pm$0.1          & 91.5$\pm$0.2  \\ \hline
			\end{tabular}
		\end{table}
	}
	Ablation studies on the hyper-parameters $\lambda_{1}$ and $\lambda_{2}$ are summarized in Table~\ref{ablation_lam}. In general, adopting asymmetric values of $\lambda_{1}$ and $\lambda_{2}$ is helpful, which allows both the federated classifier and each client's personalized classifier to be diverse and thus improves the generalization performance. 
	On the other hand, putting more reliance on training personalized classifier is beneficial, among which setting $\lambda_{2}=3$ and $\lambda_{1}=1$ brings the best performing federated model on both datasets. 
	Furthermore, explicit regularization can improve generalization through divergence reduction between the federated and each client's personalized classifier, leading to consistent performance improvement in both bACC and bAUC by up to an average increase of 1.0\%.  
	
	\subsection{Directions of Consistency Regularization in FCA}
	
	To figure out the potential impact of the direction of consistency regularization in FCA, additional ablation studies are conducted as summarized in Table~\ref{ablation_comp}. When the federated and each client's personalized classifier co-regularize each other, \eg, federated $\leftrightarrow$ personalized, their target logits predictions are inconsistent, which is detrimental. Comparatively, each client's personalized classifier acts as a more consistent guide to regularize the federated model. It is because each client's personalized classifier is locally-kept and never replaced with federated averaging.
	As a result, compared to personalized $\leftarrow$ federated, the federated model trained with personalized $\rightarrow$ federated achieves an average increase of 1.1\% and 0.6\% in bACC and bAUC on skin lesion classification while achieving an average increase of 0.6\% and 0.1\% under \textbf{Split 1} and 0.4\% and 0.1\% under \textbf{Split 2} respectively for ICH classification.
	
	\section{Conclusion}\label{conclusion}
	
	This paper highlights a challenging problem of federated learning under severe inter-client class variations, where clients exhibit different class distributions or even completely missing classes. 
	We address the issue from the lens of classifier-guided learning with the objective of learning a more robust feature extractor and propose a federated classifier anchoring framework FCA by adding a personalized classifier at each client to guide the federated feature extractor through consistency regularization.
        The robustness of consistency regularization is improved by first debiasing the federated classifier and each client's personalized classifier according to each client's class distributions.
	With multiple participants, it is important to guarantee that each client benefits from collaboration, \eg, the federated model achieving the goal of performing well not only globally over multiple clients but also locally at each client which is rarely discussed in existing studies.
	Motivated by this extended requirement, we evaluate FCA and other federated learning frameworks under a more realistic evaluation setting, where both generalization and specialization are taken into consideration. 
	Under two challenging multi-source federated long-tailed settings on skin lesion and intracranial hemorrhage classification, FCA consistently outperforms different learning frameworks in both specialization and generalization, demonstrating that FCA is a more reliable solution for practical federated medical image classification.

\end{document}